\documentclass[conference]{IEEEtran}
\IEEEoverridecommandlockouts
\usepackage{cite}
\usepackage{amsmath,amssymb,amsfonts}
\usepackage{algorithmic}
\usepackage{graphicx}
\usepackage{textcomp}
\usepackage{xcolor}
\def\BibTeX{{\rm B\kern-.05em{\sc i\kern-.025em b}\kern-.08em
    T\kern-.1667em\lower.7ex\hbox{E}\kern-.125emX}}

\usepackage{amsmath,graphicx,amssymb,bm,url,subcaption, siunitx-v2}
\usepackage{pgf,import,lmodern} 
\usepackage{multirow} 
\usepackage{balance}

\newcommand*\emailat{\includegraphics[width=6.5px]{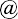}}

\def\x{{\bm x}}
\def\P{{\bm{\mathrm{P}}}}
\def\z{{\bm z}}
\def\u{{\bm u}}
\def\v{{\bm v}}
\def\F{{\bm{\mathrm{F}}}}
\def\Q{{\bm{\mathrm{Q}}}}
\def\H{{\bm{\mathrm{H}}}}
\def\R{{\bm{\mathrm{R}}}}
\def\K{{\bm{\mathrm{K}}}}
\def\I{{\bm{\mathrm{I}}}}
\def\S{{\bm{\mathrm{S}}}}

\def\p{{p}}
\def\w{{\mu}}
\def\vnu{{\bm{\mathrm{\nu}}}}

\def\sig{{\sigma}}
\def\lik{{\Lambda}}
\def\ps{{\Psi}}
\def\params{{\bm{\mathrm{\theta}}}}

\begin{document}

\title{Learning IMM Filter Parameters from\\ Measurements using Gradient Descent}

\author{\IEEEauthorblockN{{Andr\'e Brandenburger}}
	\IEEEauthorblockA{
		\textit{Fraunhofer FKIE}\\
		Wachtberg, Germany \\
		andre.brandenburger{\emailat}fkie.fraunhofer.de}
	\and
	\IEEEauthorblockN{Folker Hoffmann}
	\IEEEauthorblockA{
		\textit{Fraunhofer FKIE}\\
		Wachtberg, Germany \\
		folker.hoffmann{\emailat}fkie.fraunhofer.de}
	\and
	\IEEEauthorblockN{Alexander Charlish}
	\IEEEauthorblockA{
		\textit{Fraunhofer FKIE}\\
		Wachtberg, Germany \\
		alexander.charlish{\emailat}fkie.fraunhofer.de}
}

\maketitle

\begin{abstract}
The performance of data fusion and tracking algorithms often depends on parameters that not only describe the sensor system, but can also be task-specific. While for the sensor system tuning these variables is time-consuming and mostly requires expert knowledge, intrinsic parameters of targets under track can even be completely unobservable until the system is deployed. With state-of-the-art sensor systems growing more and more complex, the number of parameters naturally increases, necessitating the automatic optimization of the model variables. In this paper, the parameters of an interacting multiple model (IMM) filter are optimized solely using measurements, thus without necessity for any ground-truth data. The resulting method is evaluated through an ablation study on simulated data, where the trained model manages to match the performance of a filter parametrized with ground-truth values.
\end{abstract}


\section{Introduction}
\label{sec:intro}

\begin{figure}
	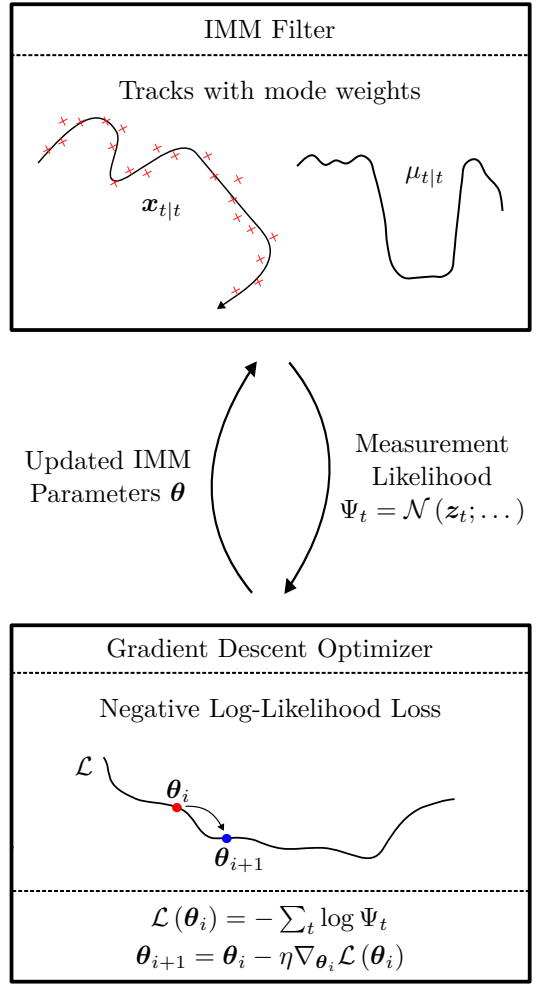
	\caption{An overview of the IMM filter optimization strategy. Based on its parametrization $\params$, the IMM filter generates tracks (top) which are utilized by the gradient descent optimizer (bottom). The optimizer successively updates the IMM filter parameters with the goal to minimize the negative log-likelihood of the measurements.}
	\label{fig:teaser}
\end{figure}

The parametrization of complex models is a prevalent topic of discussion in many research areas. While the machine-learning community already utilizes powerful automated optimization frameworks, the classical data fusion models are often parametrized manually. In this paper, we present a method to tune the popular interacting multiple model (IMM) filter\cite{bar2004estimation} solely using measurements. By defining an appropriate loss function, gradient descent can be used to update the parameters successively. A schematic overview of the algorithm is shown in Fig.~\ref{fig:teaser}.

While the IMM filter can already be seen as an approximator of target noise~\cite{Li1994, Mazor1998, Afshari2017}, the performance of the filter still strongly depends on its parametrization. Since the IMM filter allows for different noise characteristics in its modes, it is intuitively able to interpolate between the noise covariances of the modes. However, this comes with two major caveats. Firstly, the mode transition probabilities have to be set accordingly, as these are not automatically estimated, thus still requiring manual tuning. Secondly, using the IMM filter for noise estimation utilizes the mode weight as an interpolation scale, rather than an estimate of the target mode. This is an unwanted side-effect, as the mode probability is of interest in many applications, for instance when actions are triggered based on detected maneuvers.

A classical approach to filter optimization is the method of \textit{expectation maximization}, where the model is optimized by successive calculation of the expected parameter likelihood, followed by its maximization~\cite{lei2006expectation,huang2004imm,huang2005expectation}. In the literature, this optimization is realized with a potentially complex analytical calculation of the maximum, which may need to be recalculated if the underlying model or sensor architecture changes.

More recently, Barrat et al.~\cite{barratt2020fitting} have applied convex optimization tools to train Kalman smoother parameters. While it is simpler to employ, the method also requires analytical calculations, which can impede the use of more complex architectures. 

Similarly, Abbeel et al.~\cite{Abbeel2005} have shown that an extended Kalman filter (EKF) can be trained on data using coordinate ascent and Greenberg et al.~\cite{Greenberg2021} utilized gradient descent to optimize a Kalman filter. While the filter can be trained on ground-truth data~\cite{Abbeel2005, Greenberg2021}, for instance using the residual error, Abbeel et al. also present a method to optimize the measurement likelihood without need for ground-truth data~\cite{Abbeel2005}. When applied on a real system, \cite{Abbeel2005} was able to outperform hand-tuned parametrizations. 

Closely related, Xu et al.~\cite{xu2021ekfnet} presented \textit{EKFNet}, which fits the parameters of an EKF using gradient descent, analogous to the optimization of a neural network. While this approach is mostly independent to changes in the sensor system, the method utilizes parameter regularization functions, which are problem specific and need to be defined before the method can be employed.

In addition, Coskun et al.~\cite{coskun2017long} proposed an online generation of the Kalman filter matrices using long-short-term-memory (LSTM) networks. While the method outperforms traditional Kalman filters and LSTM networks, the generation of the noise covariances using neural networks can lead to an arbitrary and unexpected filter behavior. This is also the case when neural networks are directly applied to replace parts or even a complete filter~\cite{jung2019sequential, jung2020mnemonic, iter2016target}. While this is the state-of-the-art for very complex environments and dynamical latent states~\cite{doerr2018probabilistic, hafner2020dreamerv2}, the trained neural network is naturally opaque. This means, that significant efforts have to be taken to estimate the behavior of the model, which can lead to issues in explainability and can result in uncertifiability. Furthermore, these methods typically require training on ground-truth data, which is only rarely available.

In this paper, we present a method to learn parameters of an IMM filter from measurements using gradient descent. Similar to \cite{Abbeel2005}, this method utilizes a loss based on the measurement likelihood, however applied on the more difficult IMM filter. This loss function does neither necessitate regularization nor ground-truth data. The optimization of the filter is independent to the sensor system, only requiring the dynamics- and measurement model to be differentiable. The learned parameters are traditional variables of the IMM filter, which can be easily interpreted and confirmed by operators. Experiments support that the optimized IMM filter performs better than an optimized single-mode Kalman filter. In addition, it is shown that the optimized filter matches the performance of ground-truth parameters.

	\section{Interacting Multiple Model Filter}
\label{sec:method}
The Kalman filter~\cite{kalman1960new} is the foundation of many tracking algorithms. Given a target state $\x_{t}\in\mathbb{R}^n$ at time step $t$, the Kalman filter assumes linear state dynamics and a linear measurement process, both corrupted by Gaussian noise. For non-linear differentiable state dynamics, the EKF can be utilized, where the state transition is defined as
\begin{equation}
	\x_{t+1} = f(\x_{t}) + \u_t \,,\quad \u_t \sim \mathcal{N}\left(0, \Q\right)\,,
	\label{eq:ekfx}
\end{equation}
for a differentiable state transition function $f$ and Gaussian process noise $\u_t$ parametrized by the covariance matrix $\Q$. Furthermore, a measurement $\z_t\in\mathbb{R}^l$ of the true state can be observed at each time step $t$, where
\begin{equation}
	\z_t = h(\x_t) + \v_t \,,\quad \v_t \sim \mathcal{N}\left(0, \R\right)\,,
\end{equation}
for a differentiable measurement model $h$ and measurement noise $\v_t$ with corresponding covariance matrix $\R$. The construction of a Bayesian filter on these assumptions leads to the EKF equations
\begin{align}
	\F_t &= \frac{\partial f}{\partial \x}\left(\x_{t-1 \vert t-1}\right) \label{eq:F}\\
	\x_{t \vert t-1} &= f(\x_{t-1 \vert t-1}) \\
	\P_{t \vert t-1} &= \F_t \P_{t-1 \vert t-1} \F_t^T + \Q \\[2.5ex]
	\H_t &= \frac{\partial h}{\partial \x}\left(\x_{t \vert t-1}\right) \label{eq:H}\\
	\S_{t} &= \H_t \P_{t \vert t-1} \H_t^T + \R \label{eq:S}\\
	\K_{t} &= \P_{t \vert t-1}\H_t^T\S_{t}^{-1}
\end{align}
\begin{gather}
	\x_{t \vert t} = \x_{t \vert t-1} + \K_{t}\left(\z_{t}-h(\x_{t \vert t-1})\right) \\
	\P_{t \vert t} = \left(\I - \K_{t}\H_t\right)\P_{t \vert t-1}\left(\I - \K_{t}\H_t\right)^T +  \K_{t} \R \K_{t}^T\,,\label{eq:joseph_form}
\end{gather}
for a state estimate $\x$ with covariance $\P$, where the subscript $\bullet_{t \vert t}$ denotes the posterior at time $t$ and $\bullet_{t+1 \vert t}$ the prediction respectively.
Note that Eq.~\eqref{eq:joseph_form} is the \textit{Joseph form} of the covariance update which is stable with regard to suboptimal Kalman gains $\mathbf{K}_t$\cite[p.~206]{bar2004estimation}. 
This is required for iterative parameter optimization, as the Kalman gain is suboptimal per definition, due to the incorrect parameters of the filter before training.

A popular extension to the Kalman filter is the IMM filter. By defining multiple state transition models $f^{i}$ it is possible to model different dynamical modes $i=1, \dots, m$ of the system. Typically this corresponds to maneuvering modes of a target, but it can also be applied to system state estimation\cite{jo2011interacting}. At each time step the IMM filter assumes the system dynamics to be defined as
\begin{gather}
	P(i_{t+1} = j| i_{t} = i) = \p^{ij}\label{eq:modemarkov}\\
	\x_{t+1} = f^{i_t}(\x_{t}) + \u_t \,,\quad \u_t \sim \mathcal{N}\left(0, \Q^{i_t}\right)\label{eq:modeprocess}\,.
\end{gather}
Following Eq.~\eqref{eq:modemarkov}, the subsequent dynamics mode $i_{t+1}$ is sampled according to the Markov chain defined with the transition probabilities $\p^{ij}$ relative to the currently active dynamics mode $j=i_t$. Consequently, the mode transition probabilities $\p^{ij}$ define the probability of switching from mode $i$ to mode $j$ over the course of a single time step. Thus, $\p^{ij}$ satisfies $\forall i=1 \dots m: \sum_{j=1}^m \p^{ij} = 1$. Depending on the active mode, specific state transition functions $f^{i}$ are used in the process model. As shown in Eq.~\eqref{eq:modeprocess}, these different transition functions are parametrized by a mode-specific process noise $\Q^{i}$.

During the IMM filter iteration, the states of the individual modes are calculated separately through the parallel execution of $m$ differently parametrized Kalman filters. For each mode $i$, a corresponding weight $\mu^i_t$ is assigned with respect to the mode transition probabilities. The IMM filter recursion can be expressed as
\begin{gather}
	\w_{t\vert t-1}^j = \sum_{i=1}^{m} \p^{ij}\w_{t-1 \vert t-1}^i \label{eq:cbar}\\
	\w_{t-1 \vert t-1}^{i \vert j} = \p^{ij}\frac{\w_{t-1 \vert t-1}^i}{\w_{t\vert t-1}^j}  \\ 
	\x_{t-1 \vert t-1}^{0j} = \sum_{i=1}^{m} \w_{t-1 \vert t-1}^{i \vert j}\x_{t-1 \vert t-1}^i \\
	\begin{split}
		\P_{t-1 \vert t-1}^{0j} = \sum_{i=1}^{m} \w_{t-1 \vert t-1}^{i \vert j} \Big( \P_{t-1 \vert t-1}^i +
		\\ \left[ \x_{t-1 \vert t-1}^i - \x_{t-1 \vert t-1}^{0j} \right] \left[ \x_{t-1 \vert t-1}^i - \x_{t-1 \vert t-1}^{0j} \right]^T \Big)
	\end{split} \\[1.3ex]
	\lik_t^j = \mathcal{N}\left(\z_t; h(\x^{0j}_{t \vert t-1}), \S^{0j}_t\right) \label{eq:lambda} \\ 
	\w_{t \vert t}^j = \frac{1}{c} \lik_t^j\w_{t\vert t-1}^j \label{eq:wfiltered} \qquad
	c = \sum_{j=1}^{m} \lik_t^j\w_{t\vert t-1}^j\\[2.5ex]
	\x_{t \vert t} = \sum_{j=1}^{m} \w_{t \vert t}^j \x_{t \vert t}^j \label{eq:matchstate} \\ 
	\begin{split}
		\P_{t \vert t} = \sum_{j=1}^{m} \w_{t \vert t}^j \Big( \P_{t \vert t}^j + \left[\x_{t \vert t}^j - \x_{t \vert t}\right] \left[\x_{t \vert t}^j - \x_{t \vert t}\right]^T\Big)
	\end{split} \,, \label{eq:matchcov}
\end{gather}
where $\S^{0j}_t$ corresponds to the predicted innovation covariance with respect to $\x_{t-1 \vert t-1}^{0j}$ (see Eq.~\eqref{eq:S}). A detailed explanation of the IMM framework can be found in \cite[pp.~453-457]{bar2004estimation}. Note that Eq.~\eqref{eq:matchstate} and Eq.~\eqref{eq:matchcov} only have to be computed when a condensed state representation is required, but are not required for the IMM filter recursion. 

\section{IMM Parameter Optimization}
Each optimization technique requires an objective{-} or loss-function. If ground-truth data is supplied, a loss can be easily implemented through the mean-squared-error (MSE) between the ground-truth and the filter outputs. In reality, however, ground-truth data is only rarely available and the true state can only be partially observed through noisy measurements. While \cite{xu2021ekfnet} has shown that single-mode models can also be optimized with the MSE loss between the measurements and filter output, the optimization of the IMM filter is more challenging. Specifically, it has to be ensured that the training does not converge to a single-mode representation, i.e. $\forall i: \p^{ii} < 1 $ and  $\forall i \neq j: \Q^i \neq \Q^j$. 

Consequently, the loss should not only be defined on the error with respect to the state, but must also contain information about the distinct modes. In our approach, this is achieved by utilizing the measurement likelihood with respect to the predicted measurement. Namely, the filter predicts the moment-matched measurement distribution
\begin{gather}
	\hat{\x}_{t \vert t-1} = \sum_{i=1}^m \w_{t \vert t-1}^i \x_{t \vert t-1}^i \\
	\vnu_{t \vert t-1}^i = h\left( \x_{t \vert t-1}^i \right) - h\left(\hat{\x}_{t \vert t-1}\right)\\
	\hat{\S}_{t \vert t-1} = \sum_{i=1}^m  \w_{t \vert t-1}^i \left( \S_t^i + \vnu_{t \vert t-1}^i  \left(\vnu_{t \vert t-1}^i\right)^T \right) \,,
\end{gather}
where $h$ corresponds to the measurement function and $\S_t^i$ is the covariance of the expected measurement in mode $i$ (see Eq.\eqref{eq:S}). Given the model parameters $\params$ and using this distribution, the likelihood of the observed measurement is evaluated
\begin{gather}
	\ps_t\left(\params\right) = \mathcal{N}\left( \z_{t}; h\left(\hat{\x}_{t \vert t-1}\left(\params\right)\right), \hat{\S}_{t \vert t-1}\left(\params\right) \right)\,.
\end{gather}
The joint likelihood over multiple time steps can be calculated through the multiplication of the corresponding likelihoods. Since the product of probabilities can quickly approach zero, a log-space transform can be applied to keep the values numerically representable for longer time sequences and particularly suboptimal parameter initializations. This leads to a negative-log-likelihood loss
\begin{equation}
	\mathcal{L}\left(\params\right) = -\log \prod_{t} \ps_t\left(\params\right) = -\sum_{t}\log \ps_t\left(\params\right)\,.
\end{equation}

While the derivation of the loss is focused on the measurement likelihood and thus directly encodes the effects of the measurement model, the state prediction is also influenced by the mode transition and process model. Therefore, it is possible to optimize all model parameters with the single loss function $\mathcal{L}\left(\params\right)$. 

Due to its simplicity and broad applicability, gradient descent is one of the most prevalent methods for optimizing  differentiable models. This is further reinforced by the easy implementation through widely available autodifferentiation frameworks. In each training iteration $k$, the parameters $\params_k$ of the model are updated according to
\begin{equation}
	\params_{k+1} = \params_k - \eta\nabla_{\params_k}\mathcal{L}\left(\params_k\right)\,,
\end{equation}
where $\eta$ is the learning rate.

It is visible in Section \ref{sec:method} that most operations in the IMM filter are linear. In addition, the state dynamics, measurement model and the evaluation of the likelihood on a multivariate Gaussian distribution are differentiable. Thus, the gradient $\nabla_{\params}\mathcal{L}\left(\params\right)$ of the loss with respect to the model parameters $\params$ can be calculated as $\frac{\partial\mathcal{L}\left(\params\right)}{\partial\params}$ using the repeated application of the chain rule. This process can also be handled automatically by utilizing autodifferentiation. Using autodifferentiation adds to the flexibility of the method for practical operation, just as well as optimization, since the {dynamics-} and measurement model can be adapted, without requiring manual recalculation of the gradient functions.

\section{Experimental Results}
\label{sec:experiments}

\begin{figure*}
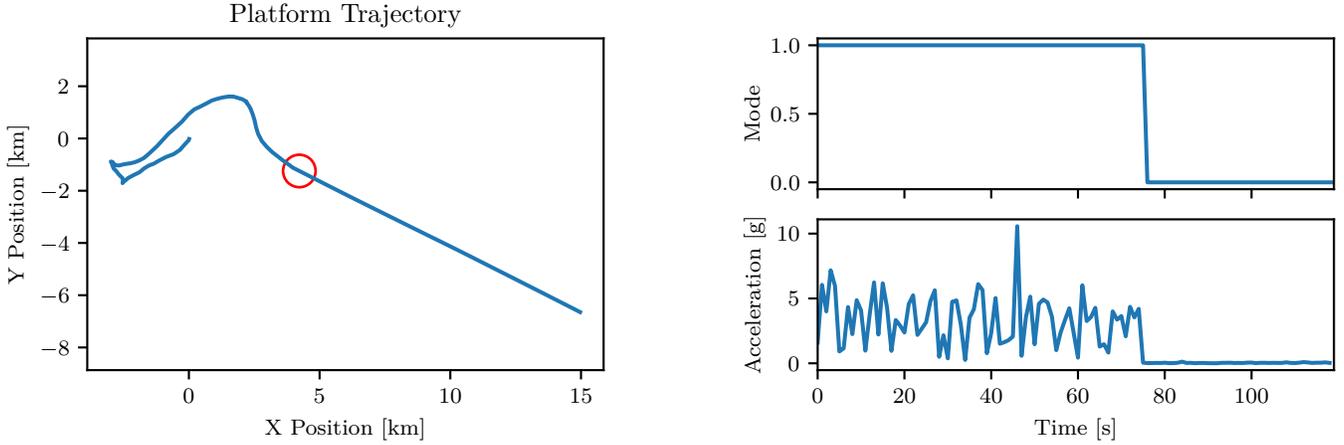

	\begin{center}
		\input{figs/trajectory.pgf}
		\hfill
		\input{figs/modes_combi.pgf}
	\end{center}
	\caption{A sample from the datasets. The ground-truth trajectory is shown on the left and starts at the coordinate origin. The mode switch is highlighted by the red circle. The ground-truth target motion mode and the absolute acceleration (process noise) are displayed on the right.}
	\label{fig:dataset}
\end{figure*}

To assess the capabilities of the method, $100$ IMM filters with linear state dynamics and measurement models are trained on simulated datasets. In addition, the quality of the loss function is evaluated and the performance of an optimized IMM filter is compared to a trained Kalman filter.

\subsection{Data Generation}

 Each of the $100$ datasets consists of $60$ trajectories, for each of which $120$ positional measurements are generated. In each dataset, the trajectories are generated with identical parameters, which are sampled uniformly at random prior to the trajectory generation. The full state vector consists of position and velocity in two dimensions. The used kinematic model is the 2-dimensional discretized white noise acceleration model\cite[p.~270]{bar2004estimation}, which is defined for each mode $i$ as
\begin{gather}
	\F = \begin{bmatrix}
		1 & \tau & 0 & 0\\
		0 & 1 & 0 & 0 \\
		0 & 0 & 1 & \tau \\
		0 & 0 & 0 & 1
	\end{bmatrix}\\
	\Q^i = \left(\sig_{v}^i\right)^2 \begin{bmatrix}
		\frac{1}{3}\tau^3 & \frac{1}{2}\tau^2 & 0 & 0\\
		\frac{1}{2}\tau^2 & \tau & 0 & 0 \\
		0 & 0 & \frac{1}{3}\tau^3 & \frac{1}{2}\tau^2 \\
		0 & 0 & \frac{1}{2}\tau^2 & \tau
	\end{bmatrix}\,.
\end{gather}
In addition, a linear measurement model is utilized, s.t. 
\begin{equation}
	\H = \begin{bmatrix}
		1 & 0 & 0 & 0\\
		0 & 0 & 1 & 0
	\end{bmatrix}\,, \quad
	\R = \begin{bmatrix}
		\sig_{r}^2 & 0 \\
		0 & \sig_{r}^2
	\end{bmatrix}\,.
\end{equation}
The initial parameters are sampled uniformly at random from the intervals
\begin{align}
	\sig_{v}^0 \in [10^{-3}, 0.98]\,, \quad
	\sig_{v}^1 & \in [9.81, 49.1]\,, 
	\label{eq:sample1}\\
	\p^{ii} \in [0.95, 0.999] \,,\hspace{2.2em} \sig_{r}& \in [1, 25]\,,
	\label{eq:sample2}
\end{align}
where $\sig_v^i$ corresponds to the process noise parameter of dynamics mode $i$ and $\sig_{r}$ sets a diagonal measurement standard deviation. The range of the process noise magnitude is set s.t. the non-maneuvering state generates peak accelerations in the order of $0.1$~g, while {$1$-$5$~g} are allowed when the target is maneuvering (see \cite[p.~270]{bar2004estimation}). The simulation step interval is $\tau=\SI{1}{\second}$ and the trajectories are generated according to Eq.\eqref{eq:modemarkov} and Eq.~\eqref{eq:modeprocess}. An example trajectory is shown in Fig.~\ref{fig:dataset}. In total, the parameters of the utilized model can be summarized as $\params = \{ \sig_{v}^0, \sig_{v}^1, \p^{00}, \p^{11}, \sig_{r} \}$. 

\subsection{Loss Evaluation}

\begin{figure}
	\begin{center}
		\input{figs/Q0_std.pgf}\\
				\vspace{-3pt}
		\input{figs/Q1_std.pgf}\\
				\vspace{-3pt}
		\input{figs/MTP0.pgf}\\
				\vspace{-3pt}
		\input{figs/MTP1.pgf}\\
				\vspace{-3pt}
		\input{figs/R_std.pgf}\\
		\vspace{-3pt}
\begingroup%
  \makeatletter%
  \providecommand\color[2][]{%
    \errmessage{(Inkscape) Color is used for the text in Inkscape, but the package 'color.sty' is not loaded}%
    \renewcommand\color[2][]{}%
  }%
  \providecommand\transparent[1]{%
    \errmessage{(Inkscape) Transparency is used (non-zero) for the text in Inkscape, but the package 'transparent.sty' is not loaded}%
    \renewcommand\transparent[1]{}%
  }%
  \providecommand\rotatebox[2]{#2}%
  \newcommand*\fsize{\dimexpr\f@size pt\relax}%
  \newcommand*\lineheight[1]{\fontsize{\fsize}{#1\fsize}\selectfont}%
  \ifx\svgwidth\undefined%
    \setlength{\unitlength}{235.44953354bp}%
    \ifx\svgscale\undefined%
      \relax%
    \else%
      \setlength{\unitlength}{\unitlength * \real{\svgscale}}%
    \fi%
  \else%
    \setlength{\unitlength}{\svgwidth}%
  \fi%
  \global\let\svgwidth\undefined%
  \global\let\svgscale\undefined%
  \makeatother%
  \begin{picture}(1,0.05449143)%
    \lineheight{1}%
    \setlength\tabcolsep{0pt}%
    \put(0.11145663,0.01415332){\makebox(0,0)[t]{\lineheight{1.25}\smash{\begin{tabular}[t]{c}\footnotesize Loss $\mathcal{L}$\end{tabular}}}}%
    \put(0,0){\includegraphics[width=\unitlength,page=1]{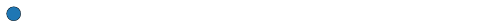}}%
    \put(0.38036976,0.01334447){\makebox(0,0)[t]{\lineheight{1.25}\smash{\begin{tabular}[t]{c}\footnotesize Posterior RMSE\end{tabular}}}}%
    \put(0,0){\includegraphics[width=\unitlength,page=2]{legend_nobox.pdf}}%
    \put(0.79583872,0.01334453){\makebox(0,0)[t]{\lineheight{1.25}\smash{\begin{tabular}[t]{c}\footnotesize Ground-Truth Parameters\end{tabular}}}}%
    \put(0,0){\includegraphics[width=\unitlength,page=3]{legend_nobox.pdf}}%
  \end{picture}%
\endgroup%

	\end{center}
	\caption{A projection of the loss for different parameters. For each graph, a single variable is assessed for a fixed dataset. The corresponding ground-truth value of the respective IMM parameter in the analyzed dataset is marked red. The loss with respect to the IMM parameter is shown in blue and the posterior RMSE to the ground-truth trajectories is shown in orange.}
	\label{fig:varsweep}
\end{figure}

To analyze the quality of the loss function, it is evaluated for different IMM parameters in proximity of the dataset parameters. In each experiment, only one variable is altered, while all other parameters are fixed to the ground-truth. Consequently, the resulting graphs in Fig.~\ref{fig:varsweep} correspond to a projection of the loss function on each parameter dimension. It is visible that the local minima of the loss function are located reasonably close to the ground-truth parameters of the dataset. Furthermore, the shown projections are smooth and mostly convex, indicating that the gradient descent optimizer is well suited. 

The local minima of the posterior RMSE with respect to the ground-truth trajectories are similarly close to the minima of the loss function, with the exception of the process noise and mode transition probability for the first mode. Here it is important to underline that this evaluation is conveyed on a projection of the loss function and thus only provides partial insight on its general shape. In particular, these projections do not capture the correlations that exist between the variables, such as the process noise and mode transition probabilities. In addition, since only a limited number of trajectories are generated, sampling error can also have an effect. Still, it seems that under-estimating the process noise is preferred for minimization of the RMSE for this dataset.

\subsection{Training}
Each dataset is split $1\mathbin{:}1$ into {train-} and test-set. Thus, the filter is trained on $1$~hour of data ($3600$ measurements) in total. The model is initialized with parameters that are drawn analogously to the dataset variables following Eq.~\eqref{eq:sample1} and Eq.~\eqref{eq:sample2}. In the following ablation study, whenever a model parameter is not trained, it is initialized with the true value that is used for dataset generation. The training is performed over $K=1000$ epochs using the \mbox{AMSGrad} optimizer\cite{reddi2019convergence} with a learning rate of $\eta=2.0\times10^{-2}$. The training takes approximately three minutes on four CPU cores with $\SI{2.6}{\giga\hertz}$ clock frequency.

\begin{table*}
	\begin{center}
		\begin{tabular}{ c c c | l | r r }
			Number of& Train $\Q^i$ & \multirow{2}{*}{Train $\R$} & \multirow{2}{*}{Metric} & Trained model & Trained model\\
			modes & and $\p^{ij}$ &  & & vs. untrained & vs. true param.\\\hline
			\multirow{4}{*}{2} & \multirow{4}{*}{Yes}& \multirow{4}{*}{Yes} & State Prediction RMSE & -16.07\% & -0.53\% \\
			& & & State Posterior RMSE & -15.30\% & -0.78\% \\
			& & & Mode Prediction MAE & -40.71\% & -6.99\% \\
			& & & Mode Posterior MAE & -44.68\% & -6.52\% \\\hline
			\multirow{4}{*}{2} & \multirow{4}{*}{Yes}& \multirow{4}{*}{No} & State Prediction RMSE & -3.67\% & -0.49\% \\
			& & & State Posterior RMSE & -3.87\% & -0.76\% \\
			& & & Mode Prediction MAE & -23.87\% & -7.29\% \\
			& & & Mode Posterior MAE & -24.50\% & -7.01\% \\\hline
			\multirow{4}{*}{2} & \multirow{4}{*}{No}& \multirow{4}{*}{Yes} & State Prediction RMSE & -13.46\% & -0.04\% \\
			& & & State Posterior RMSE & -11.42\% & -0.03\% \\
			& & & Mode Prediction MAE & -23.57\% & +0.18\% \\
			& & & Mode Posterior MAE & -28.78\% & +0.29\% \\\hline
			\multirow{2}{*}{1} & \multirow{2}{*}{Yes}& \multirow{2}{*}{Yes} & State Prediction RMSE & -8.11\% & -0.02\% \\
			& & & State Posterior RMSE & -7.16\% & +0.07\% \\\hline
			\multirow{2}{*}{1} & \multirow{2}{*}{Yes}& \multirow{2}{*}{No} & State Prediction RMSE & -3.67\% & -0.02\% \\
			& & & State Posterior RMSE & -2.70\% & +0.00\% \\\hline
			\multirow{2}{*}{1} & \multirow{2}{*}{No}& \multirow{2}{*}{Yes} & State Prediction RMSE & -7.07\% & -0.01\% \\
			& & & State Posterior RMSE & -6.37\% & +0.02\% \\
		\end{tabular}
	\end{center}
	\caption{Performance comparison of method ablations. Each ablation is evaluated on $100$ differently parametrized datasets. The entries show the change in root-mean-squared-error (RMSE) for the state and the mean-absolute-error (MAE) for the mode probability estimate. Each ablation is compared to the performance of the filter before training and to a filter that is parametrized with the true dataset variables. Negative values indicate an improvement from the baselines.}
	\label{tbl:results}
\end{table*}

Table~\ref{tbl:results} shows the method performance relative to the initial parameters and a filter based on the parameters that generated the corresponding dataset. Furthermore, an ablation study over the combinations of either training the measurement noise or process noise, just as well as a comparison to a single-mode Kalman filter is made. In all cases, the state estimation quality of the model improves compared to the initial parameters, most significantly by $16\%$ for the two-mode IMM filter where all parameters are trained. In a direct comparison, optimizing the measurement noise has a stronger impact than optimizing the process noise parameters. However, the optimization of the process model proves to be beneficial for the estimation of the mode probabilities where it enables the model to outperform the filter with ground-truth parameters by $7\%$ in estimation of the mode weights. A possible explanation for this improvement can be the dissimilarity between the modes of the optimized process noise parameters, which appears larger than in the ground-truth. Therefore, the filter may estimate the individual modes with higher confidence. In combination with the sub-optimality of the IMM filter, this can explain the improvement to the filter parametrized with true variables\cite{bar2005imm}. 

In addition, Table~\ref{tbl:results} shows that a single-mode Kalman filter can also be trained using this approach. For this setting, the datasets are restricted to a single mode. Since these experiments consider linear state dynamics and measurement models, the model with true parameters produces optimal results up to a sampling error. Remarkably, the trained filter closely matches the performance of this optimal filter.

\subsection{Comparison of IMM and Single-Mode Kalman Filter}

\begin{table}
	\begin{center}
		\begin{tabular}{ c c | l | r }
			Train $\Q^i$ & \multirow{2}{*}{Train $\R$} & \multirow{2}{*}{Metric} & IMM filter\\
			and $\p^{ij}$ &  & & vs. KF \\\hline
			\multirow{2}{*}{Yes}& \multirow{2}{*}{Yes} & State Prediction RMSE & -17.33\% \\
			& & State Posterior RMSE & -16.88\% \\\hline
			\multirow{2}{*}{Yes}& \multirow{2}{*}{No} & State Prediction RMSE & -17.34\% \\
			& & State Posterior RMSE & -16.83\% \\\hline
			\multirow{2}{*}{No}& \multirow{2}{*}{Yes} & State Prediction RMSE & -29.33\% \\
			& & State Posterior RMSE & -22.74\% \\\hline
		\end{tabular}
	\end{center}
	\caption{Direct comparison between a trained Kalman filter and the ground-truth IMM filter on targets with two motion states. The IMM filter outperforms the Kalman filter in all cases. Note for the last comparison, that the IMM filter is initialized with the true motion model, while the Kalman filter is initialized with the motion model of the first mode. Since the Kalman filter can not optimize on this initialization, the difference between the Kalman filter and IMM filter are most apparent.}
	\label{tbl:immvsekf}
\end{table}

In contrast to \cite{Abbeel2005, xu2021ekfnet}, the proposed approach utilizes an IMM filter. To assess the necessity of optimizing an IMM filter versus a Kalman filter on data that features multiple motion modes, experiments for a direct comparison have been conducted. As shown in Table~\ref{tbl:immvsekf}, the optimized IMM filter outperforms the optimized Kalman filter by more than $16\%$ in all settings. It is clearly visible, that the relative performance of the Kalman filter and the IMM filter is not affected by enabling or disabling the optimization of the measurement model. This is expected, as the measurement models of the IMM filter and Kalman filter are identical. Altogether, this indicates that the measurement models of the two filters converge similarly. The optimization without the motion model is clearly in favor of the IMM filter which outperforms the Kalman filter by more than $22\%$. While the IMM filter is initialized with the correctly parametrized two-mode motion model, it is crucial to note that the Kalman filter is initialized with a single mode corresponding to the non-maneuvering mode. Since the motion models are not optimized, the Kalman filter is not able to improve on this faulty initialization. In summary, the observations in this experiment are twofold. Firstly, the results confirm the importance of utilizing an IMM filter on data that is generated with multiple motion models. Secondly, it is shown that the optimization of a single-mode Kalman filter can help to alleviate the effects of modeling errors, such as violations of the assumptions on the target motion model.

The results of the experiments clearly show that the proposed strategy and the formulated loss function are well suited for IMM filter optimization. Furthermore, the resulting optimized filter matches the track accuracy of the ground-truth parameters. In addition, the optimized filter shows a significant improvement regarding motion mode prediction.

\section{Conclusion}
\label{sec:conclusion}
The proposed method enables the tuning of IMM filter parameters based only on measurement data, without the need for hand tuning parameters or ground-truth data. 
This is of significant practical importance due to the typical abundance of measurement data, while ground-truth data is rare at most.
Furthermore, hand tuning filter parameters can easily lead to modeling errors and degraded system performance. Through the performed experiments, a significant improvement in accuracy has been observed in comparison to mismatched (i.e. incorrect) model parameters. It is particularly encouraging that the method demonstrates a performance close to the ideal, but unrealistic case of known parameters. This result confirms the practical viability of this method for tuning IMM tracking filters. In the scope of future work, the proposed IMM optimization approach can be applied to more complex sensor architectures, where it could also alleviate sensor degradation, potentially even while the sensor is still in service.

\balance

\bibliographystyle{IEEEtran}
\bibliography{conference_101719}

\end{document}